\documentclass[11pt]{article}
\usepackage{fullpage}
\usepackage{hyperref}       
\usepackage{url}            
\usepackage{graphicx}
\usepackage{natbib}
\usepackage{xspace}
\usepackage{booktabs} 
\usepackage{caption}
\usepackage{subcaption}
\bibpunct{[}{]}{,}{n}{}{;}

\title{Tarsier: Recipes for Training and Evaluating Large\\Video Description Models}

\author{ 
Jiawei Wang\thanks{Equally contributed and are sorted in alphabetical order. $\dag$Corresponding author.} \qquad Liping Yuan$^*$ \qquad Yuchen Zhang$^{*\dag}$ \qquad Haomiao Sun\\\\
ByteDance Research\\
{
\small
\texttt{
\{wangjiawei.424,yuanliping.0o0,zhangyuchen.zyc,sunhaomiao\}@bytedance.com
}
}
}

\date{}

\newcommand{\modelname}{Tarsier\xspace}
\newcommand{\evalname}{AutoDQ\xspace}
\newcommand{\datasetname}{DREAM-1K\xspace}

\begin{document}
\maketitle

\begin{abstract}
\noindent Generating fine-grained video descriptions is a fundamental challenge in video understanding. In this work, we introduce Tarsier, a family of large-scale video-language models designed to generate high-quality video descriptions. Tarsier employs CLIP-ViT to encode frames separately and then uses an LLM to model temporal relationships. Despite its simple architecture, we demonstrate that with a meticulously designed two-stage training procedure, the Tarsier models exhibit substantially stronger video description capabilities than any existing open-source model, showing a $+51.4\%$ advantage in human side-by-side evaluation over the strongest model. Additionally, they are comparable to state-of-the-art proprietary models, with a $+12.3\%$ advantage against GPT-4V and a $-6.7\%$ disadvantage against Gemini 1.5 Pro. When upgraded to Tarsier2 by building upon SigLIP and Qwen2-7B, it further improves significantly with a $+4.8\%$ advantage against GPT-4o. Besides video description, Tarsier proves to be a versatile generalist model, achieving new state-of-the-art results across nine public benchmarks, including multi-choice VQA, open-ended VQA, and zero-shot video captioning. Our second contribution is the introduction of a new benchmark -- DREAM-1K (\url{https://tarsier-vlm.github.io/}) for evaluating video description models, consisting of a new challenging dataset featuring videos from diverse sources and varying complexity, along with an automatic method specifically designed to assess the quality of fine-grained video descriptions. We make our models and evaluation benchmark publicly available at \url{https://github.com/bytedance/tarsier}.
\end{abstract}

\section{Introduction}

Large-scale Video-Language Models (LVLMs) have made significant progress recently. These models combine visual encoders with large language models (LLMs) to achieve zero-shot video understanding capabilities. Proprietary models, such as GPT-4V~\cite{gpt4v} and Gemini~\cite{team2023gemini,reid2024gemini}, are currently considered state-of-the-art. Open-source models~\cite{lin2023video, ataallah2024minigpt4, li2023mvbench, kim2024image, xu2024pllava} have achieved high scores on public video understanding benchmarks, especially in video question answering~\cite{xu2017msrvttqa, yu2019activitynetqa, xiao2021next, li2023mvbench, mangalam2024egoschema}. However, they have not yet demonstrated performance comparable to proprietary models in challenging open-ended generative tasks.

In this paper, we focus on \emph{fine-grained video description}, a task that encapsulates some core challenges in video understanding. In this task, the model generates a long answer in free-form text to give a detailed coverage of the video content. The model must be both comprehensive and faithful, meaning that it must cover all notable events in the video without introducing hallucinations. The challenge increases when the video includes subtle movements, such as small but meaningful body actions, or quick motions. The challenge also increases as the video grows in complexity, capturing multiple events, subjects, and shots. Additionally, since the task does not have a unique answer, how to evaluate the model is a non-trivial question.

\begin{figure}[t]
    \centering
    \includegraphics[width=\textwidth]{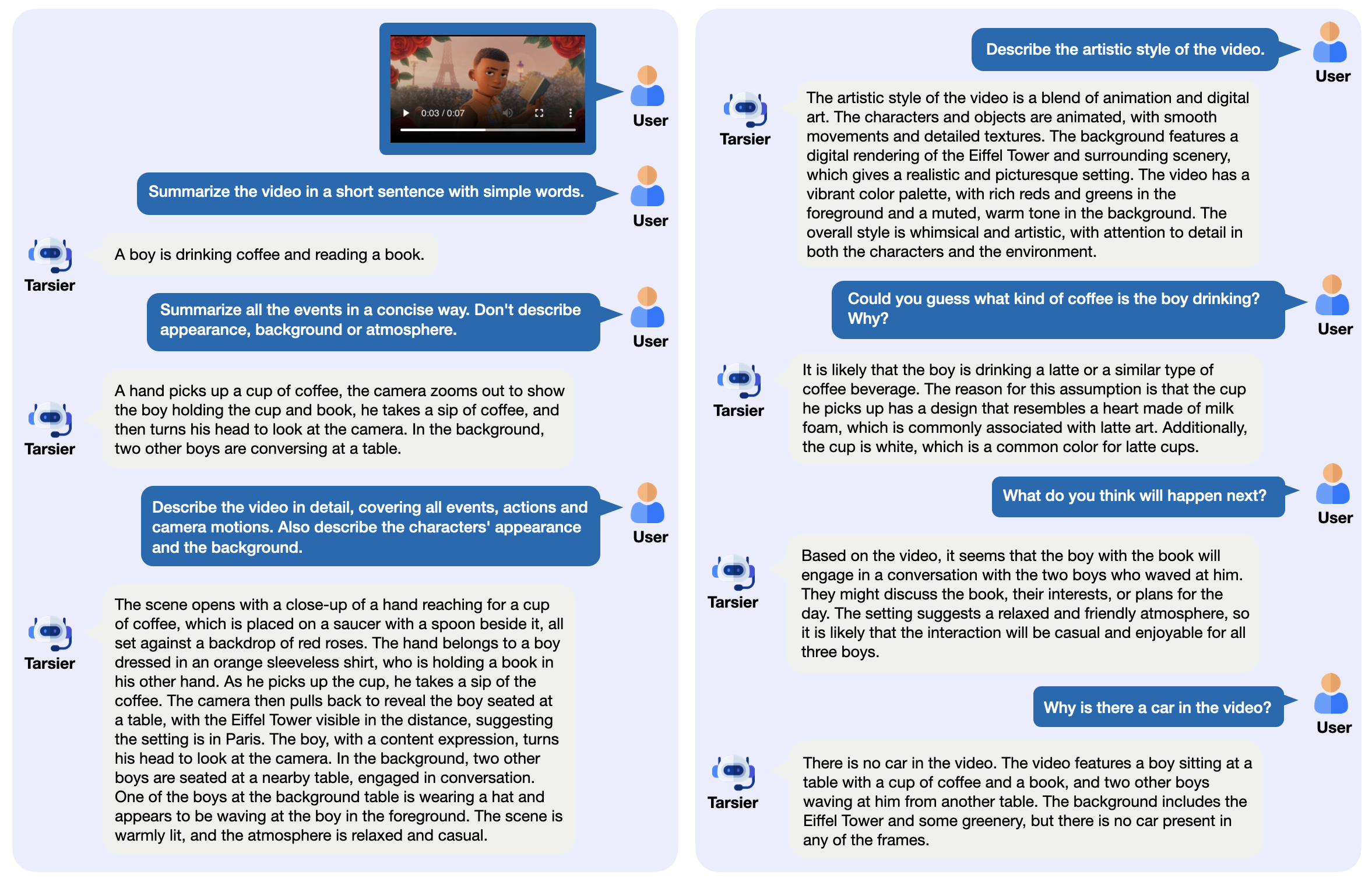}
    \caption{Example dialogue between a user and \modelname, demonstrating the model's ability to generate video descriptions with varying levels of detail. The model is also able to answer video-related questions in a zero-shot fashion. (Source movie: \emph{Turning Red, 2022})}
    \label{fig:chatbot-example}
\end{figure}

Empirically, we found that existing open source models lag far behind proprietary model in fine-grained video description. Our main contribution is the introduction of \modelname models, a family of LVLMs with simple architectures that generate high-quality video descriptions (see Figure~\ref{fig:chatbot-example} for some examples). We trained two model sizes: 7B and 34B. Each \modelname model consists of a CLIP-ViT encoder~\cite{radford2021learning}, a projection layer and a LLM. We train the model in two stages. The first stage is multi-task video-to-text pre-training, which enables the model to understand videos from different perspectives. The second stage is instruction tuning focused on generating multi-grained video descriptions. Both stages utilize high-dynamic videos with accurately matched text. Despite the simple architecture, we demonstrate that superior video description capability can be achieved by increasing the scale, diversity, and quality of the data.

Our second contribution is the introduction of a new benchmark for automatic evaluation of video description models. Existing video captioning datasets~\cite{chen2011msvd, xu2016msr, wang2019vatex} typically pair each video with a single-sentence caption. These captions are too short to stress-test the full capacity of models. We introduce a new video description dataset called \datasetname, which features videos of diverse sources and complexity. \datasetname contains 1,000 video clips selected from five categories: live-action movies, animated movies, stock videos, YouTube videos, and TikTok-style short videos. Typical \datasetname videos contain multiple events, possibly performed by multiple subjects in multiple shots, all carefully described in detail by human annotators. Additionally, we propose \evalname, an automatic evaluation method specifically designed to assess the quality of fine-grained video descriptions. \evalname breaks down the evaluation process into two steps: event extraction and entailment. This approach allows us to compute and compare the precision and recall of different models in an interpretable way.

Based on \datasetname, we demonstrate that \modelname has video description capability comparable to state-of-the-art proprietary models and significantly outperforms any existing open-source model. Specifically, the \modelname-34B model outperforms all open-source models as well as GPT-4V and Gemini 1.5 Pro in automatic evaluations. In human side-by-side evaluations, \modelname-34B significantly outperforms the strongest open-source model with a $+52\%$ advantage, is favored over GPT-4V with a $+12.3\%$ advantage, and slightly falls behind Gemini 1.5 Pro with a $-6.7\%$ disadvantage. Moreover, \modelname proves to be a versatile generalist model, achieving new state-of-the-art results on public benchmarks in multi-choice video Q\&A~\cite{li2023mvbench, xiao2021next, mangalam2024egoschema}, open-ended video Q\&A~\cite{xu2017msrvttqa, jang2017tgif, yu2019activitynetqa}, and zero-shot video captioning~\cite{chen2011msvd, xu2016msr, wang2019vatex}. Finally, we present extensive ablation studies to gain insights into what makes a good LVLM. Our findings indicate that several factors contribute to the model's strong performance: conducting large-scale multi-task pre-training, scaling up the LLM, tuning all LLM parameters, and fine-tuning the model with carefully annotated multi-grained video descriptions.

\section{Related Work}

\paragraph{Video-Language Modeling}
Video-language models typically consist of a visual encoder and a text decoder. On the encoder side, existing work either encodes video frames separately using a ViT~\cite{wang2022git, chen2023cosa, chen2024vast}, or encodes them jointly with architectures that can model spatial-temporal relationships~\cite{chen2023valor,xu2023mplug2,he2023vlab,piergiovanni2023mirasol3b,wang2024internvideo2}. For example, VALOR~\cite{chen2023valor} employs a Video Swin Transformer~\cite{liu2022video}, and Mirasol3B~\cite{wang2024internvideo2} uses TubeViT~\cite{piergiovanni2023rethinking} to model the spatial-temporal relationship, while InternVideo2~\cite{wang2024internvideo2} concatenates all the visual tokens and feeds them to a standard ViT. On the decoder side, standard Transformer~\cite{vaswani2017attention} is usually used for autoregressive generation. Typical methods for pre-training the video-language model include video-text contrastive learning~\cite{xu2021videoclip}, masked video modeling~\cite{tong2022videomae}, generative next-token prediction~\cite{alayrac2022flamingo,sun2023generative,wang2022git}, or unifying multiple training objectives~\cite{wang2023allinone,wang2022internvideo,wang2024internvideo2,zhao2024videoprism}. \modelname takes the simpler approach: it uses a frozen CLIP-ViT~\cite{radford2021learning} to encode individual frames, then rely solely on the LLM for temporal modeling. The model is trained with the next-token prediction loss only.

\paragraph{Video-LLMs} Recently, there has been a surge of research on how to connect a pre-trained video encoder to a pre-trained LLM to achieve zero-shot video understanding capabilities~\cite{li2023videochat,li2023mvbench,maaz2023video,luo2023valley,zhang2023videollama,lin2023video,ataallah2024minigpt4,wang2024elysium,xu2024pllava}. The two components are typically connected by either an adapter, such as Q-Former in BLIP2~\cite{li2023blip2}, or a simple projection layer, such as a linear layer or MLP in LLaVA~\cite{liu2023improved}. \modelname adopts the later architecture. The combined model is usually fine-tuned on a relatively small instruction tuning set. For example, Video-LLaVA~\cite{lin2023video} is fine-tuned on 100K video-text pairs, while VideoChat2~\cite{li2023mvbench} and PLLaVA~\cite{xu2024pllava} are fine-tuned on 788K video-text pairs, including 7K detailed video descriptions. Another line of research explores training-free approaches that utilize existing image understanding models or short clip captioners to achieve video understanding~\cite{zhang2023llovi,kim2024image,wang2024videoagent,kahatapitiya2024language}. In contrast, our models are pre-trained with a larger and more diverse dataset, and then fine-tuned with high-quality, fine-grained description data.

\paragraph{Video Description} Earlier research~\cite{wang2022git,xu2023mplug2,he2023vlab,chen2023valor,chen2024vast} typically pre-trains a video-language model and then fine-tunes it on specific video caption datasets, with performance measured by BLEU, ROUGE~\cite{lin2004rouge}, METEOR~\cite{banerjee2005meteor}, or CIDEr~\cite{vedantam2015cider}. Although these fine-tuned models can achieve high scores on specific datasets, they often do not generalize well to open-world videos. Some recent models report zero-shot results~\cite{alayrac2022flamingo,yan2022videococa,zhang2023videollama,zhao2024videoprism,wang2024internvideo2}. However, since the standard video captioning benchmarks~\cite{chen2011msvd,xu2016msr,wang2019vatex} are relatively simple, they do not stress-test the models' full capacity. As video complexity grows and descriptions contain more details, traditional metrics cannot accurately reflect the discrepancy between the generated descriptions and the ground truth. Maaz et al.\cite{maaz2023video} propose using ChatGPT\cite{achiam2023gpt} for automatic evaluation, which has been widely used to evaluate open-ended question answering~\cite{lin2023video,kim2024image,wang2024elysium,xu2024pllava}. We propose \datasetname and \evalname as a more challenging video description benchmark and a more interpretable method for automatic evaluation.

\section{The \modelname model} 

\subsection{Model Architecture}

\begin{figure}[t]
    \centering
    \includegraphics[width=0.8\textwidth]{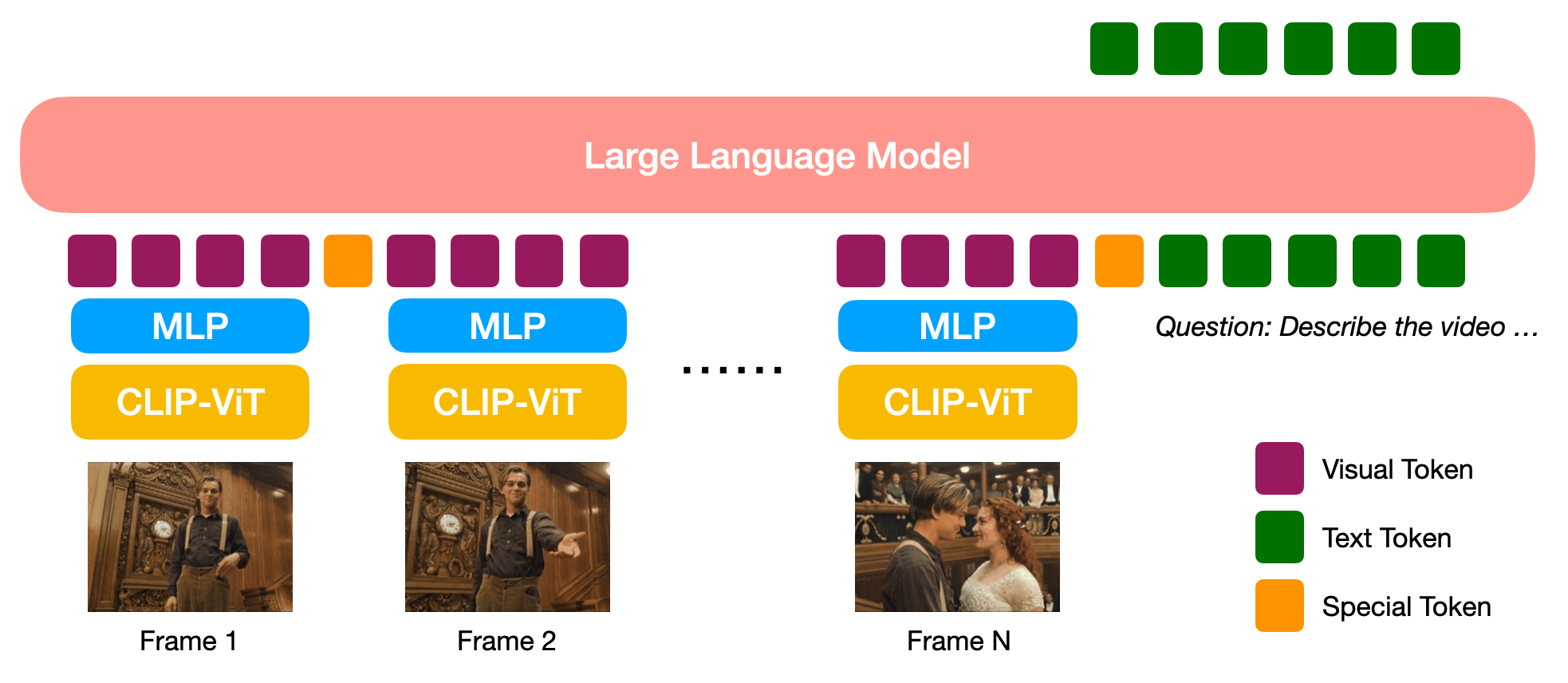}
    \caption{\modelname model architecture. We use a frozen CLIP-ViT to encode frames, then project them to the LLM token space through an MLP. The LLM is then trained with the next-token prediction loss. (Source movie: \emph{Titanic, 1997})}
    \label{fig:model-architecture}
\end{figure}

\modelname employs a simple model architecture. As depicted in Figure~\ref{fig:model-architecture}, it encodes each video frame separately. Initially, each frame is encoded using CLIP-ViT and then mapped to the downstream LLM's token embedding space via a multi-layer perceptron (MLP). The resulting visual tokens are concatenated, separated by a special \texttt{[end-of-image]} token, allowing \modelname to uniformly encode both videos and images. Then, all tokens are fed into the LLM for auto-regressive generation. This architecture captures the inter-frame relationship solely through the LLM, leveraging on its ability to do complex reasoning, which we believe is essential for understanding the connection between frames. During training, the CLIP-ViT component and the MLP are frozen, and the LLM is trained.
\subsection{Training Recipe}

We initialize \modelname with the weights from the LLaVA-NeXT image models~\citep{liu2024llavanext} and employ a two-stage training strategy\footnote{When building Tarsier2-7B, we employ SigLIP~\citep{zhai2023sigmoidlosslanguageimage} as vision encoder and Qwen2~\citep{yang2024qwen2technicalreport} as LLM. We conduct stage-0 (image-text alignment training) before the two-stage training strategy, and scale up the training data to over 38M, see https://github.com/bytedance/tarsier for more details.} Initially, we conduct multi-task pre-training using large-scale, high-quality data. Subsequently, we perform instruction tuning on moderate-scale, human annotated data. We describe the training recipe as follows; for a detailed list of all the training hyperparameters, please see Appendix~\ref{sec:training-hyperparameters}.

\subsubsection{Multi-task pre-training}

\begin{table}[t]
\centering
\small
\begin{tabular}{|lll lll|}
\hline
Dataset & Size & Task & Dataset & Size & Task \\\hline
WebVid-10M~\cite{bain2021frozen} & 2.8M & Video Captioning &
LSMDC~\cite{rohrbach2017movie} & 108K & Video Captioning \\
TGIF~\cite{li2016tgif} & 104K & Video Captioning &
ActivityNet~\cite{krishna2017dense} & 37K & Video Captioning \\
Charades~\cite{sigurdsson2016hollywood} & 16K & Video Captioning &
Charades-Ego~\cite{sigurdsson2018charades} & 6K & Video Captioning \\
YouCook2~\cite{zhou2018youcook2} & 9K & Video Captioning &
TACoS~\cite{regneri2013grounding} & 18K & Video Captioning \\
Ego4D~\cite{grauman2022ego4d} & 1.08M & Video Captioning & 
Spoken Moments~\cite{monfort2021spoken} & 492K & Video Captioning \\
TREC-VTT~\cite{awad2023trecvid} & 64K & Video Captioning & &&\\\hline 
SSV2~\cite{goyal2017something} & 168K & Action Recognition &
HMDB~\cite{kuehne2011hmdb} & 6K & Action Recognition \\
Kinetics-700~\cite{carreira2017quo} & 536K & Action Recognition &
COIN~\cite{tang2019coin} & 10K & Action Recognition \\
20BN-jester~\cite{materzynska2019jester} & 46K & Action Recognition &
Multi-Moments~\cite{monfort2021multi} & 1M & Action Recognition \\
FineAction~\cite{liu2022fineaction} & 82K & Action Recognition &
RareAct~\cite{miech2020rareact} & 2K & Action Recognition \\
DiDeMo~\cite{anne2017localizing} & 41K & Action Localization &
AVA~\cite{gu2018ava} & 27K & Action Localization \\\hline
CLEVRER~\cite{yi2019clevrer} & 83K & Video QA &
TGIF-QA~\cite{jang2017tgif} & 72K & Video QA \\
EgoQA~\cite{fan2019egovqa} & 5K & Video QA & &&
\\\hline
Oops!~\cite{epstein2020oops} & 14K & Intent Recognition &
VideoInstruct~\cite{maaz2023video} & 100K & Video Instructions \\
VIST~\cite{huang2016visual} & 38K & Visual Storytelling  & 
&& \\\hline 
ShareGPT4V~\cite{chen2023sharegpt4v} & 94K & Image Captioning &
MS COCO~\cite{lin2014microsoft} & 566K & Image Captioning \\
Flicker~\cite{plummer2015flickr30k} & 145K & Image Captioning &
LLaVA-1.5~\cite{liu2023improved} & 665K & Image Instructions \\\hline 
OpenOrca~\cite{lian2023openorca} & 1M & Text &
ShareGPT~\cite{vicuna2023} & 88K & Text \\
\hline
\end{tabular}
\caption{Public datasets used in \modelname pre-training.}
\label{tab:public-datasets}
\end{table}

In the pre-training stage, we trained our model across diverse tasks such as video captioning, video question answering, action recognition, multi-image understanding, and text generation, utilizing a broad array of public datasets as detailed in Table~\ref{tab:public-datasets}. To unify the training process, all tasks were converted into text generation tasks. For each dataset, we crafted specific prompts to define the output format. For instance, video captioning tasks may use prompts like \emph{Describe the video in one sentence} or \emph{Describe the video in detail}, tailored to the length of the annotated description; Video QA tasks may use prompts like \emph{Give a short answer to the following question}. Once prompts were assigned to each dataset, we leveraged ChatGPT~\cite{achiam2023gpt} to enhance prompt diversity. The model was subsequently trained to produce the annotated response based on the video/image input and the associated prompt. For datasets that have train/val/test splits, we use the training splits only.

Note that larger-scale video-text datasets such as HowTo100M~\cite{miech2019howto100m}, HD-VILA~\cite{xue2022hdvila}, and Panda-70M~\cite{chen2024panda} exist. We opted not to use the video-text pairs in these datasets due to issues such as narrow domains or low-quality texts. We found that human-annotated descriptions of high-dynamic videos are the most helpful for training our models. Notably, the WebVid-10M dataset~\cite{bain2021frozen} includes many instances where the video content can be described by just looking at a single frame, which are less valuable for video understanding. To address this, we used instance re-sampling to prioritize videos involving human actions and animal movements, as these are more likely to feature dynamic events. This resulted in 2.8M video-text pairs from WebVid-10M being used in pre-training.

In addition to public datasets, we also added 3.5M in-house data to pre-training, including 2.4M high-quality video caption data similar to WebVid and 1.1M videos with object-tracking annotations. The later annotations are automatically generated for a subset of videos in WebVid-10M~\cite{bain2021frozen} and HD-VILA~\cite{xue2022hdvila}. We use DEVA~\cite{cheng2023tracking} to generate object-tracking bounding boxes for these videos at high frame rates, then train the model to predict the bounding box coordinates of every object in every frame.

Overall, our pre-training dataset comprises 13.6M video-text pairs. We trained the model on 48 A100 GPUs for 100K steps using a batch size of 288. Throughout the training process, we randomly sampled between 8 to 32 frames per video. To sample $n$ frames, we first partitioned the video evenly into $n$ segments, then randomly sampled one frame from each segment. Each sampled frame was resized to square before being sent to vision encoder.

\subsubsection{Instruction Tuning}

In the instruction tuning stage, we use 500K of in-house instruction tuning data. The dataset contains a wide spectrum of video complexities and diverse prompts. In addition to the tasks from pre-training, we added the following new tasks:
\begin{itemize}
\item Describing Complex Videos at Different Levels of Granularity: We collected a set of 100K movie clips featuring multiple shots, subjects, or events, and had annotators provide descriptions varying in length and detail, from brief motion summaries to comprehensive narratives of visual details. During training, we used different prompts to precisely specify the requirements for each level of granularity.
\item Describing Camera Motions: We found that precisely describing camera motion is a weak point of existing models. To address this, we created a dataset rich in camera motions, including zooming, translating, panning, and rotating. Then we ask annotators to accurately describe these motions.
\item Creative Writing: The model was also trained to generate creative outputs based on videos, such as imagining characters' dialogues, drafting motivational speeches, and writing poems. We used self-instruct~\cite{wang2022self} to create diverse prompts and relied on GPT-4V~\cite{gpt4v} to generate responses based on human-annotated video descriptions.
\end{itemize}
By incorporating these tasks, we further enhanced the model's capabilities in multi-grained video description. We fine-tuned the model for 10K steps on the instruction tuning set using 16 A100 GPUs and a batch size of 96. During training, we sampled 16 frames from each of the 100K complex video clips mentioned above, while sampling 8 frames from other videos. This differentiation is necessary because movies often contain rapid actions and movements that can only be fully captured with a higher frame rate. The first and last frames of the video were always selected, with the intermediary frames evenly sampled throughout the video.

\section{Evaluation of Video Description Models}

In this section, we present \datasetname, a new benchmark and \evalname, an automatic evaluation method to comprehensive assess the quality of video description models.

\subsection{\datasetname: A challenging video description benchmark}

Generating high-quality video descriptions requires an understanding of multiple frames together. This understanding demands two primary capabilities. Firstly, the model should be capable of comparing frames at the pixel level to detect subtle motions, such as a person speaking or the camera panning. Secondly, it must link frames on a semantic level to reason about higher-level events, like the characters' intentions and relationships. Current video captioning benchmarks, such as MSR-VTT~\cite{xu2016msr} and VATEX~\cite{wang2019vatex}, are relatively simplistic. Typically, each video in these benchmarks involves a single major action performed by one subject in a single shot. Additionally, the \emph{single frame bias}~\cite{lei2022revealing} indicates that many actions in these datasets can be identified from still images, without the need for considering temporal relations.

\begin{table*}[ht]
\centering
\footnotesize
\begin{tabular}{l c c c c c c} 
\toprule
 & Live-action & Animation	& YouTube & Shorts & Stock & Total \\\midrule
Number of videos & 200 & 200 & 200 & 200 & 200 & 1,000 \\
Avg. video length (second) & 7.3 & 6.1 & 7.8 & 9.5 & 13.7 & 8.9 \\
Avg. text length (word) & 47.5  & 56.7 & 81.0 & 67.2 & 43.9 & 59.3 \\
Avg. number of events & 5.5 & 6.5 & 6.6 & 7.5 & 5.4 & 6.3 \\
Avg. number of subjects & 2.4 & 2.6 & 2.2 & 1.9 & 1.8 & 2.2 \\
Avg. number of shots & 2.4 & 2.1 & 2.2 & 1.5 & 1.0 & 1.9 \\\bottomrule
\end{tabular}
\caption{\datasetname data statistics.}
\label{table:dream-statistics}
\end{table*}

\begin{figure}[ht]
    \centering
    \includegraphics[width=\textwidth]{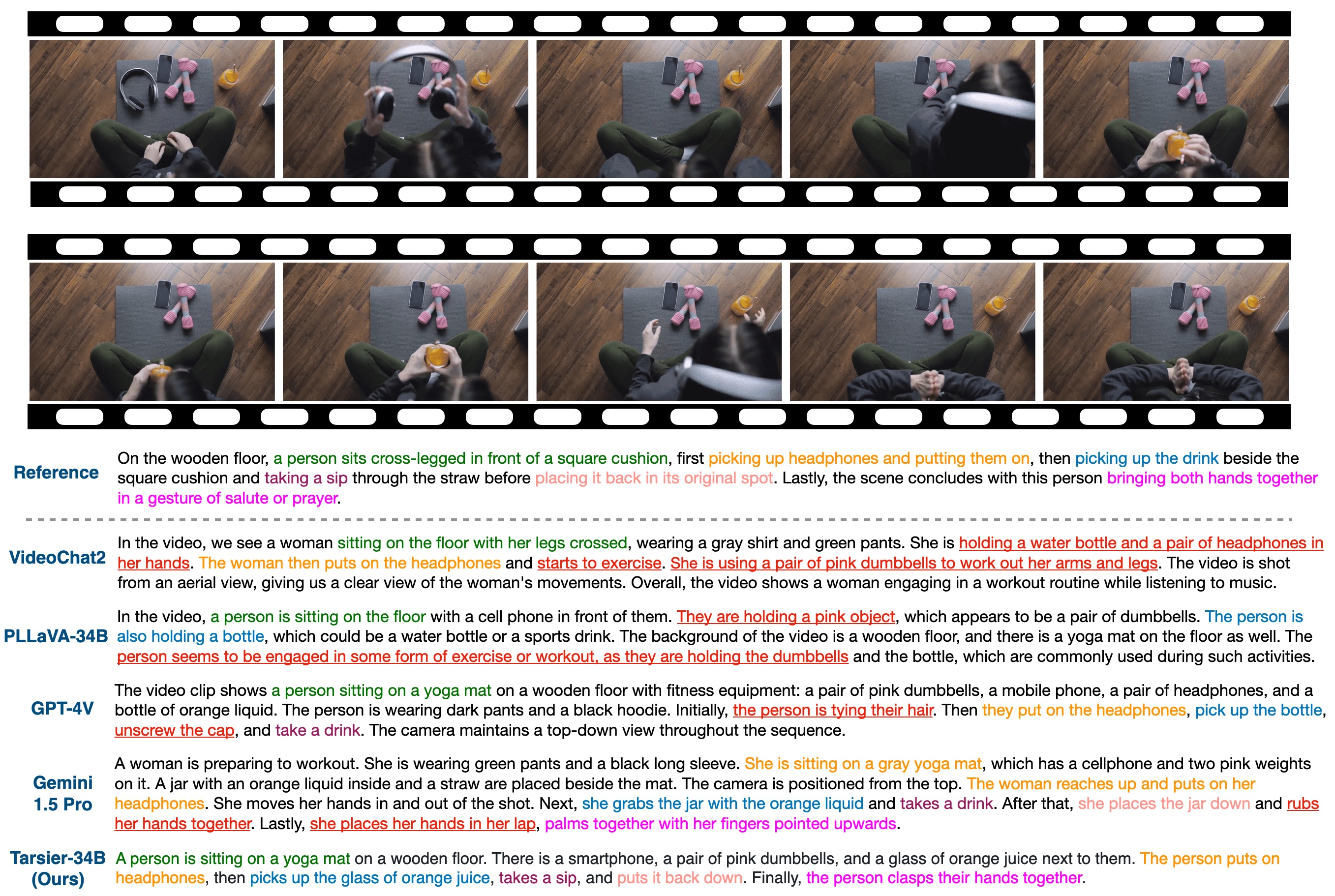}
    \caption{This video features six actions, each highlighted in a unique color. The \modelname descriptions capture more actions and contains fewer hallucinations (indicated by underlining and red color) compared to existing models.}
    \label{fig:video-description-example}
\end{figure}

To quantitatively assess the true capabilities of models, we introduce \datasetname (\textbf{D}escription with \textbf{R}ich \textbf{E}vents, \textbf{A}ctions, and \textbf{M}otions), a collection of 1,000 annotated video clips with diverse complexities. We sourced these videos from five different origins: live-action movies, animated movies, stock videos, long YouTube videos, and TikTok-style short videos, with each category contributing 200 clips. We manually pick all videos from Internet to ensure that the entire dataset is disjoint from public video-language training data. Each clip in \datasetname includes at least one dynamic event that cannot be accurately identified from a single frame alone. Table~\ref{table:dream-statistics} shows the statistics for the dataset. On average, each video lasts 8.9 seconds, includes 6.3 events, 2.2 subjects (people, animals, etc.) and 1.9 shots. We provide a fine-grained manual annotation for each video, covering all events, actions, and motions, with an average length of 59.3 words.

Figure~\ref{fig:automatic-evaluation} illustrates an example video from \datasetname. The video is 14-second long and features six actions. From this example, a noticeable performance gap emerges between models. The descriptions generated by existing models are less comprehensive, often missing key events, or less accurate, introducing hallucinations. In contrast, the descriptions generated by our \modelname model capture more events and exhibit far fewer hallucinations.

\subsection{\evalname: Automatic evaluation of description quality}
\label{sec:AutoDQ}

\begin{figure}[ht]
    \centering
    \includegraphics[width=0.9\textwidth]{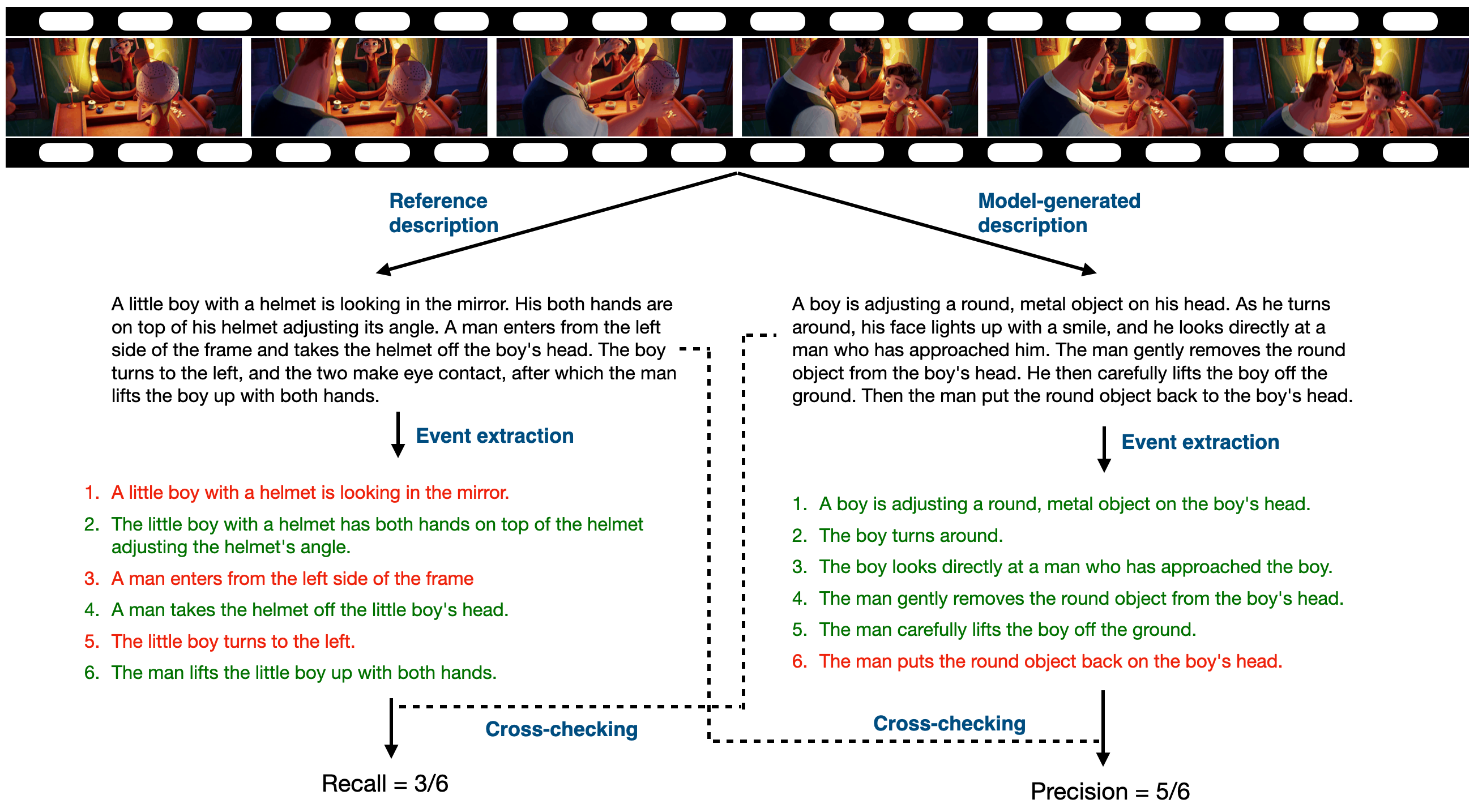}
    \caption{The \evalname workflow. \evalname uses an extraction model to extract events from two video descriptions, then uses an entailment model to examine how many events extracted from one description are entailed by the other description. We use ChatGPT to implement both models. (Source movie: \emph{Fireheart, 2022})}
    \label{fig:automatic-evaluation}
\end{figure}

Traditional $n$-gram based metrics, such as CIDEr~\cite{vedantam2015cider}, fall short when evaluating long video descriptions. This is primarily because there are numerous ways to produce a high-quality description, many of which have minimal $n$-gram overlap with the reference description. Additionally, manually assigning quality scores to descriptions is both costly and subjective. To address this, Maaz et al.~\cite{maaz2023video} propose using ChatGPT to rate descriptions on a scale from 1 to 5. However, the meaning of each rating is ambiguous, and the ratings themselves are not calibrated.

We propose \evalname (\textbf{Auto}matic \textbf{D}escription \textbf{Q}uality) as a more interpretable approach to automatic evaluation. Given a reference description $D_{\rm ref}$ and a model-generated description $D_{\rm model}$, our method compares them through a two-step procedure. In the first step, we use an \emph{event extraction model} to extract a sequence of events from both descriptions. In the second step, we employ a \emph{natural language inference model} to compute two ratios: the ratio of events in $D_{\rm ref}$ that are entailed by $D_{\rm model}$, and the ratio of events in $D_{\rm model}$ that are entailed by $D_{\rm ref}$. The first ratio is defined as the recall of $D_{\rm model}$, while the second ratio is defined as its precision. See Figure~\ref{fig:automatic-evaluation} for a concrete example. Since both event extraction and natural language inference are well-studied NLP tasks, we can rely on ChatGPT to perform them effectively. For the specific prompts used in these tasks, please refer to Appendix~\ref{sec:autodq-prompts}. Finally, by aggregating these results over all examples, we obtain the precision and recall of the model across the entire dataset, from which we can compute its F1 score.

\section{Experiments}\label{sec:experiments}

In this section, we compare the \modelname models with state-of-the-art open-source and proprietary models on a variety of video understanding tasks. 

\subsection{Results on \datasetname}\label{sec:experiments-dream-1k}

In this test, we generate detailed descriptions for each video in \datasetname. By default, for Tarsier and GPT-4V, we uniformly sample 8 frames per video. For Gemini 1.5 Pro, it dynamically samples frames at 1 FPS. For other open-source models, we sample frames following their official settings: 8 frames for Video-LLaVA; 90 frames for MiniGPT-4V; 32 frames for LLaVA-NeXT-Video; 16 frames for Tarsier2-7B, GPT-4o~\cite{gpt4o}, VideoChat2 and PLLaVA.

The default prompt is ``Describe the video in detail.'' For GPT-4V, GPT-4o and Gemini, we found that more specific instructions yield better descriptions, leading us to use a different prompt for these models. For other models, we use the default prompt unless the official implementation recommends a specific one~\cite{xu2024pllava,llavanextvideo2024}. See Appendix~\ref{sec:video-description-prompts} for the specific prompts that we have used.

\subsubsection{Automatic evaluation}

\begin{table*}[h]
\centering
\scriptsize
\resizebox{\columnwidth}{!}{
\begin{tabular}{l c c c c c c} 
\toprule
Model & Live-action & Animation	& YouTube & Shorts & Stock & Overall\\\midrule
Video-LLaVA~\cite{lin2023video} & 19.4/24.3/16.2 & 15.3/21.2/11.9 & 21.2/31.9/15.8 & 18.5/29.4/13.5 & 27.0/33.5/22.7 & 20.4/28.1/16.0
\\
MiniGPT-4V$^\dag$~\cite{ataallah2024minigpt4} & 23.6/23.8/23.4 & 16.4/19.0/14.5 & 24.8/28.4/22.1 & 23.7/29.1/20.0 & 30.7/30.3/31.1 & 24.0/26.1/22.2
\\
LLaVA-NeXT-Video$^\star$~\cite{llavanextvideo2024} & 25.8/33.4/21.0 & 22.0/32.0/16.7 & 26.5/36.2/20.9 & 24.2/39.8/17.4 & 31.1/37.3/26.7 & 26.1/35.7/20.5
\\
VideoChat2$^\ddag$~\cite{li2023mvbench} & 26.0/28.1/24.3 & 18.8/23.9/15.5 & 27.3/32.7/23.4 & 26.5/36.8/20.7 & 33.2/33.6/32.7 & 26.6/31.0/23.3
\\
PLLaVA-34B~\cite{xu2024pllava} & 29.3/34.9/25.2 & 20.9/32.0/15.6 & 28.9/40.8/22.3 & 25.6/41.9/18.4 & 35.1/42.5/29.9 & 28.2/38.4/22.3
\\\midrule
GPT-4V~\cite{gpt4v} & 34.8/39.2/31.3 & 27.4/31.9/24.0 & 33.8/40.1/29.2 & 34.8/46.1/28.0 & 40.7/46.7/36.1 & 34.4/40.8/29.7
\\
Gemini 1.5 Pro~\cite{reid2024gemini} & 36.4/36.4/36.4 & 30.7/31.8/29.7 & 34.0/36.7/31.6 & 37.0/42.4/32.7 & 42.2/40.7/43.8 & 36.2/37.6/34.8
\\
GPT-4o~\cite{gpt4o} & 39.8/42.1/37.8 & 35.8/\textbf{39.1}/33.1 & 33.8/40.1/29.2 & 39.9/\textbf{47.9}/34.2 & \textbf{44.0}/\textbf{46.6}/41.7 & 39.2/\textbf{43.4}/35.7
\\\midrule
\modelname-7B & 36.6/38.5/34.8 & 29.3/34.6/25.5 & 33.0/39.2/28.4 & 33.6/44.6/26.9 & 39.6/44.7/35.5 & 34.6/40.3/30.2
\\
\modelname-34B & 38.5/39.6/37.5 & 32.2/35.8/29.2 & \textbf{34.5}/\textbf{41.1}/29.7 & 34.0/44.1/27.7 & 41.7/46.4/37.8 & 36.3/41.4/32.4 \\
Tarsier2-7B & \textbf{43.8}/\textbf{43.1}/\textbf{44.5} & \textbf{37.1}/37.2/\textbf{37.1} & \textbf{34.5}/37.7/\textbf{31.8} & \textbf{40.9}/45.5/\textbf{37.1} & 43.9/44.2/\textbf{43.6} & \textbf{40.1}/41.5/\textbf{38.8}

\\\bottomrule
\end{tabular}
}
\caption{\datasetname evaluation by \evalname. We report F1/Precision/Recall scores for each category and for the overall dataset. \dag: \href{https://huggingface.co/Vision-CAIR/MiniGPT4-Video/resolve/main/checkpoints/video_mistral_checkpoint_last.pth?download=true}{MiniGPT-4V-Mistral}. \ddag: \href{https://huggingface.co/OpenGVLab/VideoChat2_stage3_Mistral_7B/resolve/main/videochat2_mistral_7b_stage3.pth?download=true}{VideoChat2-Mistral}. $\star$: \href{https://huggingface.co/lmms-lab/LLaVA-NeXT-Video-34B-DPO}{LLaVA-NeXT-Video-34B-DPO}. The \modelname models achieve significantly higher scores than all open source models in all categories.}
\label{table:dream-results}
\end{table*}

\begin{figure}[h]
    \centering
    \includegraphics[width=0.32\textwidth]{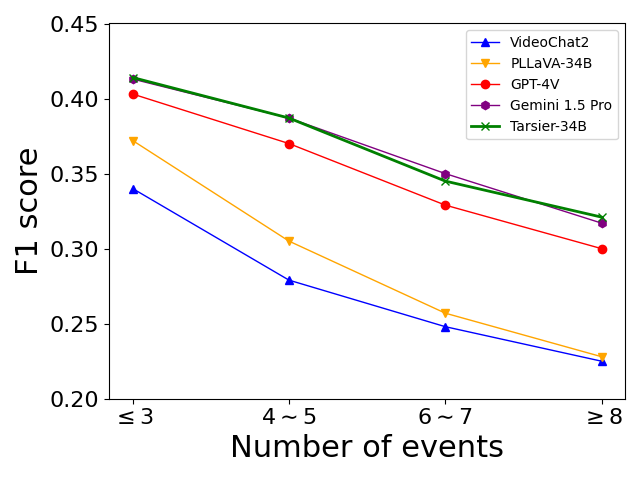}
    \includegraphics[width=0.32\textwidth]{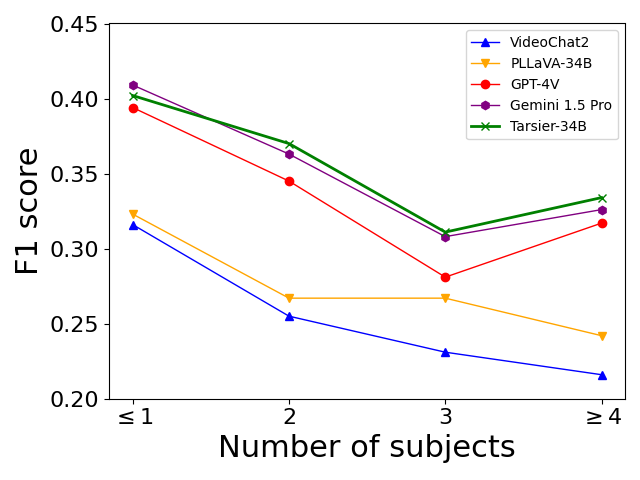}
    \includegraphics[width=0.32\textwidth]{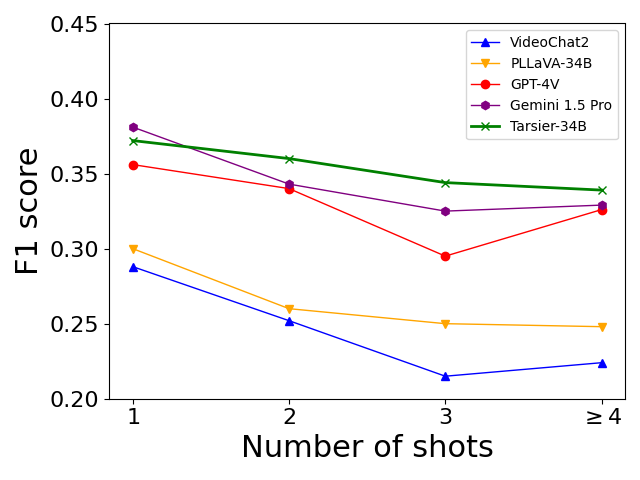}
    \caption{Description quality (F1 score) as a function of increasing video complexity.}
    \label{fig:f1-score-by-video-complexity}
\end{figure}

First, we report the automatic evaluation results computed by the method in Section~\ref{sec:AutoDQ} in Table~\ref{table:dream-results}. See the \href{https://tarsier-vlm.github.io/#leaderboard}{leaderboard} for evaluation results of more latest VLMs. The ChatGPT model used to implement \evalname is \texttt{gpt-3.5-turbo-0125}. As shown in Table~\ref{table:dream-results}, the \modelname models outperform all open-source models in terms of precision, recall, and F1 scores across all categories. Among the existing open-source models, the PLLaVA-34B model exhibits the best overall description quality, yet there is a large gap between its scores and those of \modelname models. Both the \modelname-7B and \modelname-34B models achieve higher F1 scores than GPT-4V. The 34B model also outperforms Gemini 1.5 Pro in the overall score, though the 7B model is slightly weaker than Gemini. This is a promising result, especially considering that our training procedure utilizes significantly less compute resources than these proprietary models. When built upon SigLIP and Qwen2, with a 5x samller model scale, Tarsier2-7B performs much better than Tarsier-34B by gaining a 40.1 overall F1 score, which even slightly better than the best proprietary model GPT-4o.

Next, we partitioned \datasetname according to the number of events, subjects, and shots to test the models against increasing video complexities. As shown in Figure~\ref{fig:f1-score-by-video-complexity}, while F1 scores generally decline for all models as video complexity increases, Tarsier-34B consistently outperforms other open-source models and GPT4-V across all levels of complexity, while achieving comparable scores consistently with Gemini 1.5 Pro.

\subsubsection{Human evaluation}

\begin{table*}
\centering
\small
\begin{tabular}{l c c c c} 
\toprule
Comparison & \modelname wins & Same &  Baseline wins & Advantage\\\midrule
\modelname-34B vs PLLaVA-34B & 71.7\% & 8.0\% & 20.3\% & $+51.4\%$\\
\modelname-34B vs GPT-4V & 50.0\% & 12.3\% & 37.7\% & $+12.3\%$\\
\modelname-34B vs Gemini 1.5 Pro & 28.0\% & 37.3\% & 34.7\% & $-6.7\%$\\
Tarsier2-7B vs GPT-4o & 23.2\% & 58.4\% & 18.4\% & $+4.8\%$\\\bottomrule
\end{tabular}
\caption{Side-by-side human evaluation results on \datasetname. The advantage rate is defined as the win rate subtracted by the loss rate.}
\label{table:human-evaluation}
\end{table*}

Human side-by-side comparison is a gold standard for evaluating video descriptions. In this test, we randomly sampled 300 videos from \datasetname, then asked experienced annotators to compare the descriptions generated by two different models, collecting their preferences. Every pair of descriptions was randomly shuffled to ensure that the annotators were blind to the corresponding models. We compared the \modelname-34B model against PLLaVA-34B, GPT4-V, and Gemini 1.5 Pro. The results, as shown in Table~\ref{table:human-evaluation}, align with our automatic evaluation findings. Our model significantly outperforms PLLaVA, demonstrating strong capabilities in describing dynamic events with less hallucination. It is also preferred over GPT-4V with a notable margin. When compared to Gemini 1.5 Pro, \modelname has a slightly negative advantage rate, but in a significant percentage of cases (37.3\%), the two models are labeled as ``Same''. For Tarsier2-7B, it gains a slight advantage (4.8\%) over GPT-4o, and ties with GPT-4o on over one half cases. This indicates that their performance is comparable in many instances. See Appendix~\ref{sec:human-evaluation-details} for more information about the human evaluation process.

\subsection{Results on other public benchmarks}

\begin{table*}[t]
\centering
\scriptsize

\begin{subtable}[t]{0.49\textwidth}
\begin{tabular}{l c c c} 
\toprule
Model & MVBench & Next-QA & EgoSchema \\\midrule
ST-LLM~\cite{liu2024st} & 54.9 & - & -\\
PLLaVA~\cite{xu2024pllava} & 58.1 &  - & - \\
VideoChat2~\cite{li2023mvbench} & 60.4 & - & 63.6/54.4 \\
InternVideo2~\cite{wang2024internvideo2} & 60.9 & - & - \\
IG-VLM~\cite{kim2024image} & - & 70.9 & - \\
LLoVi~\cite{zhang2023llovi} & - & 67.7 & 57.6/50.3 \\
VideoAgent~\cite{wang2024videoagent} & - & 71.3 & 60.2/54.1\\
LangRepo~\cite{kahatapitiya2024language} & - & 60.9 & 66.2/41.2\\
VideoTree~\cite{wang2024videotree} & - & 73.5 & 66.2/61.1\\
\midrule
\modelname-7B & 62.6 & 71.6 & 56.0/49.9 \\
\modelname-34B & \textbf{67.6} & \textbf{79.2} & \textbf{68.6}/\textbf{61.7} \\\bottomrule
\end{tabular}
\caption{Accuracy of zero-shot multi-choice video question answering. On EgoSchema, we report the results on subset/fullset respectively.}
\label{tab:multi-choice-vqa}
\end{subtable}
\hfill
\begin{subtable}[t]{0.49\textwidth}
\begin{tabular}{l c c c} 
\toprule
Model & MSVD & MSR-VTT & VATEX \\\midrule
BLIP2~\cite{li2023blip2} & 27.4 & 19.0 & - \\
Panda~\cite{chen2024panda} & 49.2 & 31.5 & - \\
VideoCoca~\cite{yan2022videococa} & - & 27.1 & 22.8 \\
Video-LLaMA~\cite{zhang2023videollama} & 47.0 & 29.1 & -  \\
Flamingo~\cite{alayrac2022flamingo} & - & - & 39.5 \\
VideoPrism~\cite{zhao2024videoprism} & - & 38.5 & 31.7  \\
InternVideo2~\cite{wang2024internvideo2} & 93.1 & 43.5 & 49.2 \\\midrule
\modelname-7B &  98.0 & 24.1 & 51.1   \\
\modelname-34B & \textbf{125.9} & \textbf{44.7} & \textbf{57.1} \\\bottomrule
\end{tabular}
\caption{The CIDEr scores ($\uparrow$) of zero-shot video captioning.}
\label{tab:video-captioning}
\end{subtable}
\\\vspace{20pt}

\begin{subtable}[t]{0.8\textwidth}
\centering
\begin{tabular}{l c c c c}
\toprule
Model & MSVD-QA & MSR-VTT-QA & ActivityNet-QA & TGIF-QA \\\midrule
Video-LLaVA~\cite{lin2023video} & 70.7/3.9 & 59.2/3.5 & 45.3/3.3 & 70.0/4.0 \\
MiniGPT-4V~\cite{ataallah2024minigpt4} & 73.9/4.1 & 58.3/3.5 & 44.3/3.4 & 72.2/4.1 \\
VideoChat2~\cite{li2023mvbench} &  70.0/3.9 & 54.1/3.3 & 49.1/3.3 & -\\
ST-LLM~\cite{liu2024st} & 74.6/3.9 & 63.2/3.4 & 50.9/3.3 & - \\
IG-VLM LLaVA-34B~\cite{kim2024image} & 79.6/4.1 & 62.4/3.5 & 58.4/3.5 & 79.1/4.2  \\
IG-VLM GPT-4V~\cite{kim2024image} & 76.3/4.0 & 63.8/3.5 & 57.0/3.5 & 65.3/3.7 \\
Elysium~\cite{wang2024elysium} &75.8/3.7& 67.5/3.2& 43.4/2.9 &66.6/3.6 \\
PLLaVA-7B~\cite{xu2024pllava} & 76.6/4.1 & 62.0/3.5 & 56.3/3.5 & 77.5/4.1 \\
PLLaVA-34B~\cite{xu2024pllava} & 79.9/4.2 & \textbf{68.7/3.8} & 60.9/3.7 & 80.6/4.3  \\\midrule
\modelname-7B & 77.0/4.1 & 62.0/3.5 & 59.5/3.6 &	79.2/4.2	 \\
\modelname-34B & \textbf{80.3/4.2} & 66.4/3.7 & \textbf{61.6/3.7} & \textbf{82.5/4.4}	 \\\bottomrule
\end{tabular}
\caption{Accuracy and quality score of zero-shot open-ended video question answering. We employ ChatGPT to evaluate the scores.}
\label{tab:open-ended-vqa}
\end{subtable}
\caption{Zero-shot results on public benchmarks. When citing a baseline model that has multiple variants (e.g., different sizes), we always report the results of the best-performing variant.}
\label{table:benchmark-results}
\end{table*}

We evaluated models on popular video understanding benchmarks. As \modelname is a generalist model, we report its zero-shot performance, meaning that the training sets of these datasets (if they exist) were not used in any stage of training. The results for existing models were sourced directly from their respective papers. \modelname models sampled 8 frames per video, except for the long-form video understanding benchmark, EgoSchema, where we sampled 16 frames per video.

\subsubsection{Multi-choice video question answering}

In multi-choice video question answering, the model is given a video, a question, and several answer choices, and is asked to select the correct answer. We compare models on MVBench~\cite{li2023mvbench}, Next-QA~\cite{xiao2021next}, and EgoSchema~\cite{mangalam2024egoschema}. MVBench is a comprehensive video understanding benchmark covering 20 tasks; Next-QA is a benchmark for causal and temporal action reasoning; EgoSchema is a benchmark for evaluating long video understanding capabilities. To prevent data leakage, we have removed all EgoSchema-related videos from Ego4D~\cite{grauman2022ego4d} during training.

Results presented in Table~\ref{tab:multi-choice-vqa} demonstrate that the \modelname models have established new state-of-the-art performance across three benchmarks. We observe a consistent increase in accuracy as the size of the LLM increases, which we attribute to the enhanced reasoning capabilities of larger LLMs.

\subsubsection{Open-ended video question answering}

In open-choice video question answering, the model is provided with a video and a question and is asked to generate a free-form answer. We report the results on four standard benchmarks: MSVD-QA~\cite{xu2017msrvttqa}, MSR-VTT-QA~\cite{xu2017msrvttqa}, ActivityNet-QA~\cite{yu2019activitynetqa}, and TGIF-QA~\cite{jang2017tgif}. Following the approach of Video-ChatGPT~\cite{maaz2023video}, we utilize ChatGPT to assess performance, using the same model version  (\texttt{gpt-3.5-turbo-0125}) and the same evaluation prompt as previous work to ensure comparability of results. 

The results presented in Table~\ref{tab:open-ended-vqa} show that \modelname-34B outperforms all existing models on MSVD-QA, ActivityNET-QA, and TGIF-QA, establishing new state-of-the-art results. Additionally, \modelname-7B outperforms all existing models on these datasets, with the exception of PLLaVA-34B. Similar to multi-choice VQA, we observe a consistent increase in accuracy and quality as the size of the LLM increases.

\subsubsection{Zero-shot video captioning}

In this experiment, we evaluated our model on MSVD~\cite{chen2011msvd}, MSR-VTT~\cite{xu2016msr}, and VATEX~\cite{wang2019vatex} video captioning benchmarks. Since our model was not exposed to the training sets of these benchmarks, we compared our results with previous zero-shot video captioning models. Table~\ref{tab:video-captioning} demonstrates that our \modelname models established new state-of-the-art for zero-shot video captioning. It is important to note that CIDEr scores for zero-shot evaluation are typically much lower than those for models specifically fine-tuned on these datasets. However, fine-tuned models are optimized to match specific benchmarks in terms of caption length, word usage, and language style, which can boost the CIDEr scores but does not necessarily reflect the true description quality.

\subsection{Ablation studies}

\begin{table*}[h]
\centering
\footnotesize
\begin{tabular}{l | l l | c c} 
\toprule
Model & Original setting & Ablating setting & \datasetname & MVBench \\\midrule
\modelname-7B & - & - & 34.6 & 62.6 \\\midrule
w/o pre-training & stage1+stage2 & stage2 only & 30.6 & 46.6 \\\midrule
w/o instruction tuning & stage1+stage2 & stage1 only & 25.8 & 60.2 \\\midrule
w/o multi-task pre-train & stage1: multi-task data & stage1: WebVid-10M & 31.4 & 48.4 \\\midrule
w/o multi-grained & stage2: multi-grained & stage2: VideoChat2~\cite{li2023mvbench} & 22.4 & 61.0 \\
description data & description data & training data &&\\\midrule
w/o full LLM tuning & train all LLM layers & train top-8 LLM layers & 29.9 & 49.3 \\\bottomrule
\end{tabular}
\caption{Ablation study results. We report the F1 scores on \datasetname and the accuracy scores on MVBench.}
\label{table:ablation-study}
\end{table*}

In this section, we conduct ablation studies to assess the effectiveness of our training methodology. We examine the impact of five different ablation settings: (1) omitting pre-training, (2) omitting instruction tuning, (3) substituting our 13.6M multi-task pre-training data with the WebVid-10M~\cite{bain2021frozen} video captioning data, (4) substituting our 500K instruction tuning data with the VideoChat2~\cite{li2023mvbench} training data, which contains 788K video instruction tuning pairs with only 7K detailed video descriptions, and (5) rather than tuning all parameters of the LLM, we train only its top-8 layers to reduce computational costs. Due to the high computational costs involved, our experiments are conducted on the 7B model.

Table~\ref{table:ablation-study} presents the results of our ablation studies, revealing that each modification leads to a significant decline in performance across both \datasetname and MVBench. Specifically, removing stage 1 pre-training or replacing the pre-training dataset with WebVid-10M results in a substantial drop in MVBench accuracy, with decreases of $-16.0$ and $-14.2$ absolute points respectively. The table also shows a notable drop in the \datasetname scores. These results underscore the importance of multi-task pre-training in developing a robust generalist model. On the other hand, omitting instruction tuning or changing the instruction tuning data significantly diminishes the \datasetname score by $-8.8$ and $-12.2$ absolute points, highlighting the critical role of our human-annotated, multi-grained video description data in enhancing the model's video description capabilities. Furthermore, training only the top-8 layers of the LLM leads to inferior outcomes on both datasets, emphasizing the importance of full model tuning.

\section{Conclusion and Future Work}

In this paper, we release the \modelname family of models, which outperforms existing open-source video description models in both automatic and human evaluations. Our ablation studies reveal that the strong performance of \modelname can be attributed to its extensive multi-task pre-training, as well as its fine-tuning on human-annotated, multi-grained video description data. Looking ahead, several directions for further exploration appear promising. First, there is potential for scaling up the pre-training data to the next level, which would require innovative methods to acquire video-text pairs from the internet. Second, scaling up both the visual encoder and the LLM simultaneously could yield improvements. Third, refining the model's ability to follow complex instructions could further enhance its utility.

\bibliographystyle{plain}
\bibliography{paper}

\newpage
\appendix

\section{Training hyperparameters}\label{sec:training-hyperparameters}

The following table shows the training hyperparameters in Stage 1 and Stage 2.
\begin{table}[h]
    \centering
    \begin{tabular}{l c c}
         \toprule
         Configuration            & Stage 1 (Pre-training) & Stage 2 (Instruction Tuning) \\\midrule
         ViT init.                & \multicolumn{2}{c}{\texttt{clip-vit-large-patch14-336}} \\
         LLM init.                & \multicolumn{2}{c}{LLaVA-NeXT} \\
         VL Adapter init.         & \multicolumn{2}{c}{LLaVA-NeXT} \\
         Optimizer name           & AdamW & AdamW \\
         Optimizer $\beta_1$      & $0.9$ & $0.9$ \\
         Optimizer $\beta_2$      & $0.999$ & $0.999$ \\
         Optimizer eps            & $1e^{-6}$ & $1e^{-6}$\\
         Learning rate            & $2e^{-5}$ & $2e^{-5}$ \\
         Learning rate schedule   & cosine & cosine \\
         Training steps           & 100,000 & 10,000 \\
         Warm-up steps            & 1,000 & 1,500 \\
         Weight decay             & 0.01 & 0.01 \\
         Gradient clip            & 1.0 & 1.0 \\
         Dropout rate             & 0 & 0 \\
         Global batch size        & 288 & 96 \\
         Frame resolution         & 336x336 & 336x336 \\
         Tokens per frame         & 576 & 576 \\
         Frames per video         & [8,32] & \{8,16\} \\
         Numerical precision      & \texttt{bfloat16} & \texttt{bfloat16} \\
         \bottomrule
    \end{tabular}
    \caption{Training hyperparameters of \modelname}
    \label{tab:hyperparam}
\end{table}

\section{\evalname prompts}
\label{sec:autodq-prompts}
This appendix presents the ChatGPT prompts used by \evalname to execute automatic evaluation. Below is the prompt for event extraction. There is a placeholder to put in the video description text from which events will be extracted:\\\\
\fbox{\texttt{
\begin{minipage}{\textwidth}
Below is a description of a video clip:\\
{[The video description]}\\\\
Extract at most 10 key events from the above video description paragraph. Requirements:\\
- An event must include an action, motion or movement (NOT STATIC INFOMATION). DON'T repeat same events.\\
- Every event is represented by a brief sentence with in 10 words, with a subject, a predicate and optionally an object, avoid unnecessary appearance descriptions.\\
- Every event must be atomic, meaning that it cannot be further split into multiple events.\\
- Scene cuts and camera motions are NOT events.\\
- Substitute pronouns by the nouns they refer to.\\\\
Please generate the response in the form of a Python dictionary string with keys "events". The value of "events" is a List(str), of which each item is an event. DO NOT PROVIDE ANY OTHER OUTPUT TEXT OR EXPLANATION. Only provide the Python dictionary string. For example, your response should look like this:\{"events": [event1, event2, ...]\}
\end{minipage}
}}\\\\

\noindent Below is the ChatGPT prompt for predicting entailment. There are placeholders to put in the video description text, as well as the events extracted by the previous prompt:\\\\
\fbox{\texttt{
\begin{minipage}{\textwidth}
Given a video description and a list of events. For each event, classify the relationship between the video description and the event into three classes: entailment, neutral, contradiction. \\
- "entailment" means that the video description entails the event.\\
- "contradiction" means that some detail in the video description contradicts with the event.\\
- "neutral" means that the relationship is neither "entailment" or "contradiction".\\\\
Output a list in Json format:\\
{[}
\{"event": "copy an event here", "relationship": "put class name here", "reason": "give a reason"\},
...
{]}\\\\
Video description:\\
{[The video description]}\\\\
Events:\\
{[The events in Json list-of-string format]}\\\\
DO NOT PROVIDE ANY OTHER OUTPUT TEXT OR EXPLANATION. Only output the JSON. Output:
\end{minipage}
}
}\\

\section{Video description prompts}\label{sec:video-description-prompts}

This appendix presents the prompts used for the \datasetname experiments in Section~\ref{sec:experiments-dream-1k}. The default prompt is:\\
\fbox{\texttt{
\begin{minipage}{\textwidth}
Describe the video in detail.
\end{minipage}
}}\\\\

\noindent The prompt for GPT-4V:\\
\fbox{\texttt{
\begin{minipage}{\textwidth}
Given 8 frames uniformly sampled from a video clip, describe the video (not the individual images!) in detail, focusing on the main subjects, their actions and the background scene. DON'T describe feelings or atmosphere.
\end{minipage}
}}\\\\

\noindent The prompt for Gemini 1.5 Pro:\\
\fbox{\texttt{
\begin{minipage}{\textwidth}
Describe the video in one paragraph, mainly focusing on the dynamic events in the video. Don't describe feelings or atmosphere.
\end{minipage}
}}\\\\

\noindent The prompt for PLLaVA-34B, as recommended by~\cite{xu2024pllava}:\\
\fbox{\texttt{
\begin{minipage}{\textwidth}
You are to assist me in accomplishing a task about the input video. Reply to me with a precise yet detailed response. For how you would succeed in the recaptioning task, read the following Instructions section and Then, make your response with a elaborate paragraph.\\\\
\# Instructions\\
1. Avoid providing over detailed information such as color, counts of any objects as you are terrible regarding observing these details\\
2. Instead, you should carefully go over the provided video and reason about key information about the overall video\\
3. If you are not sure about something, do not include it in you response.\\\\
\# Task\\
Describe the background, characters and the actions in the provided video.
\end{minipage}
}}\\\\

\noindent The prompt for LLaVA-NeXT-Video, as recommended by~\cite{llavanextvideo2024}:\\
\fbox{\texttt{
\begin{minipage}{\textwidth}
Please provide a detailed description of the video, focusing on the main subjects, their actions, and the background scenes.
\end{minipage}
}}\\

\section{Human evaluation details}
\label{sec:human-evaluation-details}

In the human evaluation, we randomly sampled a subset of \datasetname, consisting of 300 videos. Three experienced annotators were involved in the evaluation. For each model pair, each annotator was presented with 100 video description pairs. Each pair contained two descriptions generated by the two models, but their order was randomly swapped to ensure that the annotator could not identify the model that produced each description. The annotators were asked to assign one of three labels to each pair: ``Description A is better,'' ``Description B is better,'' or ``Same quality.'' Before the annotation process began, we provided them with the following annotation guide:

\begin{quote}
Imagine you are reading video descriptions to a blind person. Which version of the description better helps the blind person understand the content of the video? You should choose the more helpful version. This judgment may be subjective, but in principle, you can consider both accuracy and comprehensiveness:
\begin{itemize}
\item When equally accurate, the more comprehensive description is more helpful.
\item When equally comprehensive, the more accurate description is more helpful.
\item If both descriptions have flaws in accuracy and comprehensiveness, you need to judge which flaw has a lesser impact on understanding the main content of the video. In this case, you can use your own subjective feelings and common sense to make a judgment, without relying on explicit rules.
\item We value the description of dynamic events. Therefore, your evaluation should focus on dynamic events (actions, behaviors, changes, etc.). Detailed descriptions of static aspects (appearance, color, texture, static background) do not earn extra points, unless they help understand key events. The model does not accept audio input, so you can completely ignore the sounds in the video.
\item Please make an effort to discern the better description, and try to use the "Same quality" option as little as possible.
\end{itemize}
\end{quote}

\end{document}